\documentclass{article}

\usepackage{microtype}
\usepackage{graphicx}
\usepackage{subfigure}
\usepackage{booktabs} 

\usepackage{hyperref}

\usepackage[accepted]{icml2025}

\usepackage{amsmath}
\usepackage{amssymb}
\usepackage{mathtools}
\usepackage{amsthm}

\usepackage[capitalize,noabbrev]{cleveref}

\theoremstyle{plain}

\theoremstyle{definition}

\theoremstyle{remark}

\usepackage[textsize=tiny]{todonotes}

\usepackage[utf8]{inputenc} 
\usepackage[T1]{fontenc}    
\usepackage{hyperref}       
\usepackage{url}            
\usepackage{booktabs}       
\usepackage{amsfonts}       
\usepackage{nicefrac}       
\usepackage{microtype}      
\usepackage{xcolor}         

\usepackage{graphicx}  
\usepackage{subcaption}
\usepackage{svg}

\usepackage[T3,T1]{fontenc}
\DeclareSymbolFont{tipa}{T3}{cmr}{m}{n}
\DeclareMathAccent{\invbreve}{\mathalpha}{tipa}{16}

\usepackage{wrapfig}
\usepackage{booktabs}
\usepackage{siunitx}
\usepackage{colortbl}
\usepackage{hyperref}
\usepackage{url}

\icmltitlerunning{A Practical Two-Stage Recipe for Mathematical LLMs}

\begin{document}

\twocolumn[
\icmltitle{A Practical Two-Stage Recipe for Mathematical LLMs: \\Maximizing Accuracy with SFT and Efficiency with Reinforcement Learning}




\begin{icmlauthorlist}
\icmlauthor{Hiroshi Yoshihara}{aff1,aff2}
\icmlauthor{Taiki Yamaguchi}{aff3}
\icmlauthor{Yuichi Inoue}{aff4}
\end{icmlauthorlist}

\icmlaffiliation{aff1}{Aillis Inc., Tokyo, Japan}
\icmlaffiliation{aff2}{Department of Health Policy and Public Health, Graduate School of Pharmaceutical Sciences, The University of Tokyo, Tokyo, Japan}
\icmlaffiliation{aff3}{Rist Inc., Kyoto, Japan}
\icmlaffiliation{aff4}{Sakana AI, Tokyo, Japan}

\icmlcorrespondingauthor{Yuichi Inoue}{y.inoue@sakana.ai}

\icmlkeywords{Machine Learning, ICML}

\vskip 0.3in
]



\printAffiliationsAndNotice{}  

\begin{abstract}
Enhancing the mathematical reasoning of Large Language Models (LLMs) is a pivotal challenge in advancing AI capabilities. While Supervised Fine-Tuning (SFT) and Reinforcement Learning (RL) are the dominant training paradigms, a systematic methodology for combining them to maximize both accuracy and efficiency remains largely unexplored. This paper introduces a practical and effective training recipe that strategically integrates extended SFT with RL from online inference (GRPO). We posit that these methods play complementary, not competing, roles: a prolonged SFT phase first pushes the model's accuracy to its limits, after which a GRPO phase dramatically improves token efficiency while preserving this peak performance. Our experiments reveal that extending SFT for as many as 10 epochs is crucial for performance breakthroughs, and that the primary role of GRPO in this framework is to optimize solution length. The efficacy of our recipe is rigorously validated through top-tier performance on challenging benchmarks, including a high rank among over $2,200$ teams in the strictly leak-free AI Mathematical Olympiad (AIMO). This work provides the community with a battle-tested blueprint for developing state-of-the-art mathematical reasoners that are both exceptionally accurate and practically efficient. To ensure full reproducibility and empower future research, we will open-source our entire framework, including all code, model checkpoints, and training configurations at \href{https://github.com/analokmaus/kaggle-aimo2-fast-math-r1}{https://github.com/analokmaus/kaggle-aimo2-fast-math-r1}.
\end{abstract}

\section{Introduction}
\label{introduction}

The remarkable advancements of Large Language Models (LLMs) have demonstrated their potential across a vast spectrum of applications. Beyond their well-established capabilities in natural language understanding and conversation, the frontier of AI research is increasingly focused on enhancing their reasoning abilities, which are essential to solve complex and challenging problems~\cite{qwq-32b-preview, jaech2024openai, openai2025competitiveprogramminglargereasoning, deepseekai2025deepseekr1incentivizingreasoningcapability}. Among the diverse domains for evaluating reasoning, mathematical problem-solving stands out as an ideal testbed. It demands not only factual knowledge but also a complex interplay of strategic planning, step-by-step logical inference, and self-correction, thereby providing a comprehensive benchmark for the problem-solving capabilities of current models.

To enhance the mathematical reasoning of LLMs, two primary paradigms have been explored: Supervised Fine-Tuning (SFT) and Reinforcement Learning (RL). SFT, which uses high-quality step-by-step solutions datasets, has been instrumental in the bootstrapping of model capabilities~\cite{chen2024alphamath, qwq-32b-preview}. However, its effectiveness is intrinsically tied to the scale and quality of the demonstration data, potentially leading to a performance plateau. On the other hand, RL offers a promising avenue for models to learn from their own generated solutions, moving beyond the confines of static datasets. Recently, Group Relative Policy Optimization (GRPO) has emerged, showing promise in improving sampling efficiency~\cite{shao2024deepseekmath, deepseekai2025deepseekr1incentivizingreasoningcapability}. Yet, it remains unclear whether GRPO alone is sufficient to maximize performance, and a principled, systematic methodology for combining it with SFT is notably absent.

This paper bridges this gap by proposing a practical recipe that strategically integrates SFT and GRPO to unlock new levels of mathematical reasoning in LLMs. We posit that these two methods are not competing but rather play highly complementary roles. Our core idea is that an extended SFT phase is first employed to establish a strong performance baseline, which is then refined by a GRPO phase focused on enhancing efficiency without compromising accuracy. This sequential and synergistic approach provides a clear, reproducible pathway for developing high-performing mathematical reasoners.

Our investigation yields several key insights into the effective training of mathematical LLMs. First, we experimentally demonstrate that prolonged SFT is crucial for performance breakthroughs. Contrary to the common practice of short-duration fine-tuning such as cold start, we find that although initial epochs may show a temporary dip in performance, extending the SFT process for as long as 10 epochs consistently and significantly boosts the model's problem-solving accuracy. Second, we uncover a new primary role for GRPO in this combined framework. While prior work often associates preference optimization with direct accuracy gains, our findings indicate that GRPO excels at dramatically improving token efficiency. After our intensive SFT stage, GRPO maintains or slightly improves the high accuracy achieved, while substantially shortening the length of generated solutions. Third, we establish a clear synergistic relationship: SFT is responsible for pushing the performance ceiling of the model, while GRPO is responsible for optimizing the solution generation process, making the high-performing model more practical and efficient for real-world applications.

To validate the efficacy of our proposed recipe, we conduct extensive evaluations on a suite of challenging benchmarks, including AIME 2024 and AIME 2025. Crucially, we also test our model on the AI Mathematical Olympiad (AIMO)~\cite{aimo2}, one of the most competitive benchmarks with stringent measures against data leakage. Our method achieves top-tier performance, demonstrating its practical effectiveness and robustness in a highly competitive setting. These results not only underscore the power of our combined SFT-GRPO strategy but also offer the community a valuable, battle-tested recipe for advancing the frontiers of mathematical reasoning with LLMs.

To support the reproducibility of our work, we will release key artifacts at \href{https://github.com/analokmaus/kaggle-aimo2-fast-math-r1}{https://github.com/analokmaus/kaggle-aimo2-fast-math-r1}. The release will include our final model weights, the complete source code for the SFT and GRPO training and evaluation procedures, all curated datasets, and the full set of checkpoints from the GRPO stage. We believe these resources will enable the community to rigorously verify and build upon our proposed training methodology.

\begin{table*}[t]
  \centering
  \small
  \caption{\textbf{Performance and the mean token usage across model sizes for AIME 2024 and 2025.} \textit{Acc.} represents mean Pass@1(\%). \textit{Original} means Deepseek-R1-Distill-Qwen 14B.}
  \begin{tabular}{l*{6}{rr}}
    \toprule
    & \multicolumn{6}{c}{\textbf{AIME 2024}} & \multicolumn{6}{c}{\textbf{AIME 2025}}\\
    \cmidrule(lr){2-7}\cmidrule(lr){8-13}
    & \multicolumn{2}{c}{1.5B} & \multicolumn{2}{c}{7B} & \multicolumn{2}{c}{14B}
    & \multicolumn{2}{c}{1.5B} & \multicolumn{2}{c}{7B} & \multicolumn{2}{c}{14B}\\
    \cmidrule(lr){2-3}\cmidrule(lr){4-5}\cmidrule(lr){6-7}
    \cmidrule(lr){8-9}\cmidrule(lr){10-11}\cmidrule(lr){12-13}
    Method & Acc. & Tokens & Acc. & Tokens & Acc. & Tokens
           & Acc. & Tokens & Acc. & Tokens & Acc. & Tokens\\
    \midrule
    Original                   & 27.8 & 11,235 & 51.0 & 10,136 & 63.3 & 9,590
                               & 22.3 & 11,154 & 38.1 & 10,612 & 46.7 & 10,602\\
    Original + RL              & \textbf{29.5} & \textbf{10,702} & 53.6 & 7,735  & 60.9 & 7,255
                               & 23.2 & 10,354 & 39.0 & 8,176  & 41.8 & 8,246\\
    \midrule
    + SFT (10 epochs)            & 26.0 & 13,014 & 52.0 & 10,647 & 65.2 & 10,268
                               & 22.1 & 12,826 & 38.3 & 11,507 & 49.7 & 11,264\\
    \rowcolor{gray!20}
    + SFT (10 epochs) + RL       & 27.3 & 11,767 & \textbf{54.7} & \textbf{9,577}  & \textbf{66.0} & \textbf{7,932}
                               & \textbf{23.2} & \textbf{11,480} & \textbf{39.8} & \textbf{10,445} & \textbf{49.2} & \textbf{9,066}\\
    \bottomrule
  \end{tabular}
  \label{tab:aime-results}
\end{table*}

\section{Related Work}
\label{related}

\paragraph{LLM Reasoning.}

The performance of Large Language Models (LLMs) is well-known to improve with the scaling of training compute budgets, as established by scaling laws \citep{kaplan2020scaling, hoffmann2022trainingcomputeoptimallargelanguage}. More recently, a parallel research direction has focused on increasing the computational budget at inference time, a strategy often referred to as test-time scaling. Several studies have demonstrated that allocating more compute during the inference phase—for instance, by generating a larger number of tokens or candidate solutions—can significantly enhance LLM performance on complex reasoning tasks \citep{li2022competition, lewkowycz2022solving, brown2024large, wu2025inference, misaki2025wider}. This line of work underscores the value of trading inference-time resources for higher accuracy.

\paragraph{Post-training for Reasoning LLMs.}

A significant body of work has focused on enhancing the reasoning abilities of LLMs through post-training refinement \citep{qwq-32b-preview, openai2025competitiveprogramminglargereasoning, deepseekai2025deepseekr1incentivizingreasoningcapability, ye2025limoreasoning, muennighoff2025s1simpletesttimescaling, kimiteam2025kimik15scalingreinforcement}. Methods such as those used for OpenAI's o1 \citep{jaech2024openai} and o3 \citep{openai2025competitiveprogramminglargereasoning}, and DeepSeek-R1 \citep{deepseekai2025deepseekr1incentivizingreasoningcapability}, fine-tune models using supervised learning and/or reinforcement learning. This trend has been mirrored in the open-source community, with a growing number of studies reporting substantial improvements in reasoning by post-training models~\citep{chen2024alphamath, openr1, zeng2025simplerl, wen2025lightr1curriculumsftdpo, dang2025reinforcement}. Despite this progress, a definitive and principled training recipe has yet to emerge, leaving the optimal combination and scheduling of techniques an open question. Furthermore, much of the existing work has centered on maximizing accuracy, often relying on increased token generation at inference time as a primary driver of performance. Consequently, a practical training methodology that explicitly considers both accuracy and inference efficiency remains a critical, yet largely underexplored, area of research.

\paragraph{Efficient Reasoning.}

While the aforementioned studies focus on improving reasoning accuracy, a parallel line of inquiry addresses the challenge of reasoning efficiency. The advent of Chain-of-Thought (CoT) prompting~\citep{wei2022chain}, despite its success, often leads to the "overthinking" phenomenon, where models generate excessively verbose rationales, incurring substantial computational overhead and latency. To mitigate this, one major research direction has focused on augmenting the model's intrinsic ability to reason concisely through training-centric paradigms. The first approach involves Reinforcement Learning (RL), where reward functions are explicitly designed to penalize generation length, thereby guiding the model to discover shorter yet effective reasoning paths~\citep{luo2025o1pruner, yeo2025demystifying, aggarwal2025l1}. A second approach leverages Supervised Fine-Tuning (SFT) with curated, variable-length CoT data. Such datasets are typically created either by post-processing and compressing verbose reasoning traces from a teacher model~\citep{kang2025c3ot, xia2025tokenskip} or by prompting models to generate shorter solutions during the data collection phase itself~\citep{liu2024can, munkhbat2025self}. Our work is situated within this context but offers a distinct, synergistic perspective. While prior works often treat accuracy and efficiency as a direct trade-off to be optimized jointly, our recipe decouples these objectives. We first employ an extended SFT phase dedicated to maximizing problem-solving accuracy, and subsequently use GRPO not primarily for further accuracy gains, but to refine the high-performing model to be significantly more token-efficient.

\section{Methods}
\label{methods}

Our training methodology proceeds in two principal stages. The first stage involves intensive Supervised Fine-Tuning (SFT) using a specially curated dataset of high-difficulty mathematical reasoning problems. This is followed by a second stage where GRPO is applied to enhance the model's token efficiency while preserving its reasoning accuracy.

\subsection{Stage 1: Supervised Fine-Tuning (SFT)}

The SFT dataset was meticulously constructed by amalgamating data from three distinct sources. From OpenR1 Math~\cite{openr1}, we selected approximately 3,000 examples where the reference DeepSeek-R1-Distill-Qwen model's solution trace was notably long, exceeding 12,800 tokens, and where its accuracy was above 50\%. An additional 3,000 examples were included from the same source where the R1 model's accuracy fell between 50\% and 75\%. The openr1 hard dataset contributed around 2,500 challenging samples sourced from \texttt{open-r1-math-220k}~\cite{openr1}; these were problems that the DeepSeek-R1-Distill-Qwen-32B model was unable to solve in four attempts. Finally, we incorporated the second-stage SFT data from the Light-R1-SFT Data~\cite{wen2025lightr1curriculumsftdpo}.

After merging these sources, we removed duplicate entries. For each unique problem, we selected the correct generation that exhibited the shortest token length. In instances where samples from Light-R1-SFT Data lacked ground truth answers, we extracted and substituted the answers from the R1 model's solution traces. This comprehensive process yielded a high-difficulty dataset consisting of 7,900 problem-solution trace-answer triplets.

For the SFT process, full-parameter SFT was performed, executed on a system equipped with 8 NVIDIA H200 GPUs. The training was configured with a per device train batch size of 1 and gradient accumulation steps of 8. The training process was extended to 10 epochs. The maximum sequence length was set to 24,000, and packing was enabled. A learning rate of 1e-5 was used with a cosine learning rate scheduler. The models were trained using the system prompt: "Please reason step by step, and put your final answer within \textbackslash boxed{\{\{\}\}}".

\subsection{Stage 2: GRPO for Enhanced Token Efficiency}

While the SFT stage improved the model's accuracy, it also led to a tendency to generate longer, sometimes redundant, reasoning traces. To mitigate this and specifically enhance token efficiency, we subsequently applied GRPO.

The dataset employed for GRPO training was the second-stage data from Light-R1-SFT Data~\cite{wen2025lightr1curriculumsftdpo}, maintaining consistency with one of the SFT data sources. The SFT-tuned model resulting from Stage 1 served as the initialization point for making GRPO phase stable.
The reward function for GRPO was designed with three key components to guide the learning process. Firstly, a Format Reward provided a binary signal (+1 or 0) based on adherence to the expected output structure, particularly matching the regular expression pattern \verb|r"^.*?oxed\{(.*?)\}.*?</think>.*?$"|. Secondly, a Cosine Similarity Reward was implemented~\cite{yeo2025demystifying}. For outputs that conformed to the required format, this reward measured the cosine similarity between the embedding of the model's generated reasoning trace and that of a reference correct trace, where available, or used a proxy based on answer correctness. This reward was scaled to range from 0.1 to 1.0 for correct answers, thereby more subtly penalizing longer correct traces, and from -1.0 to -0.1 for incorrect answers, which more severely penalized shorter incorrect traces. The maximum trace length considered for this reward was 30,000 tokens. This component offers a more nuanced, continuous feedback signal compared to a simple accuracy-based reward. Thirdly, a Length Penalty was applied, proportional to the length of the generated output, to explicitly discourage overly verbose solutions and promote conciseness.

The GRPO training was configured with num generations set to 8 and a beta value of 0.04. The per device train batch size was 2, with gradient accumulation steps of 8. This stage was conducted for 50 steps, using a maximum completion length of 16,384. The learning rate was 4e-6, again with a cosine scheduler. The system prompt was: "You are a helpful and harmless assistant. You are Qwen developed by Alibaba. You should think step-by-step. Return final answer within \textbackslash boxed\{\{\}\}".

\section{Experiments}
\label{experiments}

In this section, we present a series of experiments designed to rigorously evaluate our proposed training recipe, which was developed with the primary goal of achieving high performance on the AI Mathematical Olympiad (AIMO). We first describe our experimental setup, including the models and benchmarks used for evaluation. We then analyze the impact of our combined SFT and RL approach on both accuracy and efficiency across standard benchmarks. Finally, we report the performance of our best model on AIMO, validating the effectiveness of our recipe in the highly competitive environment.

\subsection{Benchmarks}

A significant challenge in evaluating LLMs is the rapid overfitting of models to publicly available benchmarks. To address this, the AI Mathematical Olympiad (AIMO) is structured as a competition with strictly controlled test data~\cite{aimo2}. It provides a dedicated evaluation environment during the competition period, ensuring that the benchmark remains entirely leak-free as participants have no access to the test cases. AIMO uses a public set of 50 problems for in-competition performance monitoring and a private set of 50 problems for the final performance assessment after the competition concludes. In this paper, we evaluate our models using data from the second AIMO competition. In addition, to evaluate the performance and efficiency of our method, we used the competition-level benchmarks AIME 2024 and AIME 2025, and the standard mathematical reasoning benchmark MATH-500~\cite{hendrycks2021measuring}. Throughout the following sections, we report model performance as pass@1, averaged over 64 sampling runs.

\begin{table}[t]
  \centering
  \small
  \caption{\textbf{Mean Pass@1 (\%) and mean output token length on \textsc{MATH500}.}}
  \label{tab:pass1_tokenlen}
  {%
    \setlength{\tabcolsep}{2pt}
    \begin{tabular}{l*{6}{r}}
      \toprule
      & \multicolumn{6}{c}{\textbf{MATH500}}\\
      \cmidrule(lr){2-7}
      & \multicolumn{2}{c}{1.5B} & \multicolumn{2}{c}{7B} & \multicolumn{2}{c}{14B}\\
      \cmidrule(lr){2-3}\cmidrule(lr){4-5}\cmidrule(lr){6-7}
      \textbf{Method} & Acc. & Tokens & Acc. & Tokens & Acc. & Tokens\\
      \midrule
      Original                     & 78.2 & 1,915 & 83.8 & 2,382 & 86.4 & 2,556 \\
      + SFT (10 epochs)            & 79.6 & 2,450 & 84.4 & 3,082 & 88.1 & 2,969 \\
      + SFT (10 epochs) + RL       & \textbf{83.9} & \textbf{2,210} & \textbf{89.0} & \textbf{2,849} & \textbf{91.2} & \textbf{2,084} \\
      \bottomrule
    \end{tabular}%
  }
\end{table}

\subsection{Models}

Due to the constraints of the AIMO competition, where many teams utilized 14B-parameter models, we also focused on maximizing the performance of a model of this size. To investigate how our training recipe generalizes to other model scales, we also experimented with 1.5B and 7B parameter models. For our base models, we used the DeepSeek-R1-Distill-Qwen models~\cite{deepseekai2025deepseekr1incentivizingreasoningcapability}, which were also a popular choice among AIMO participants.

\begin{figure}[t]
\begin{center}
\centerline{\includegraphics[width=\columnwidth]{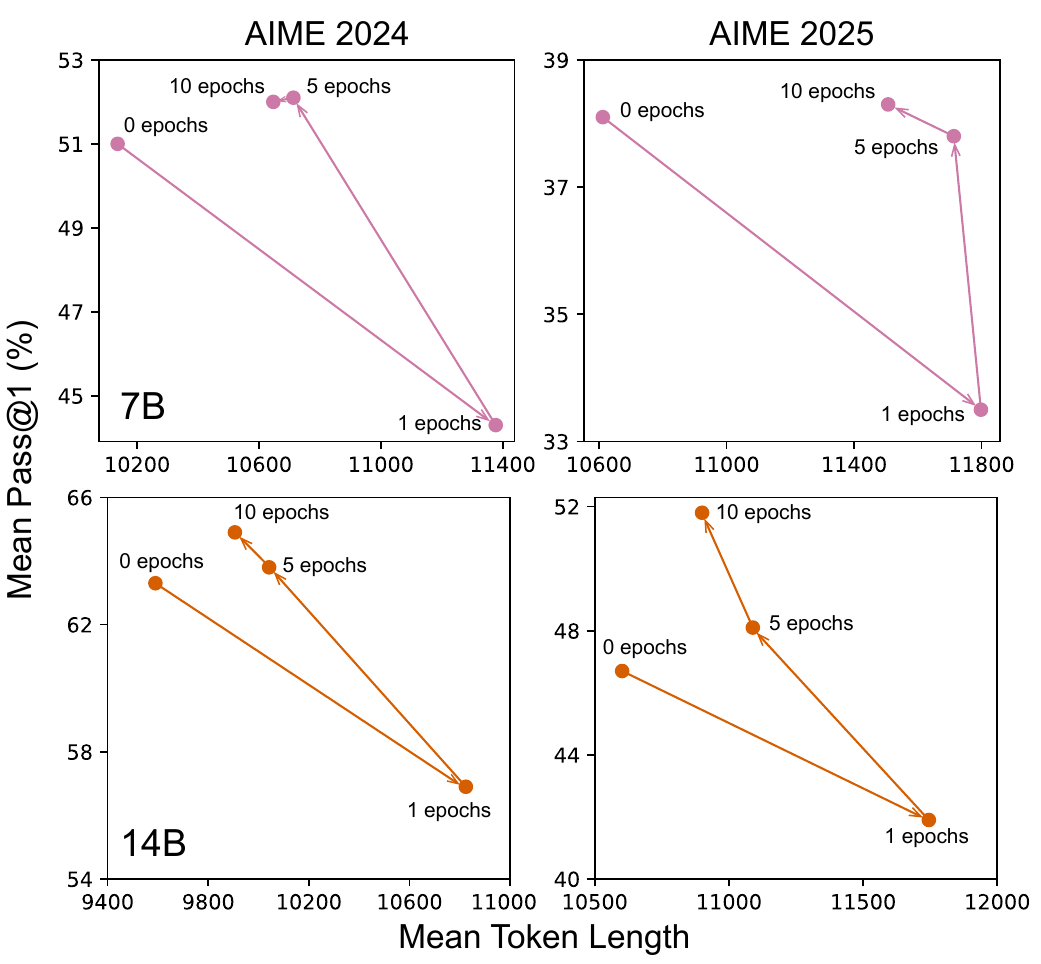}}
\caption{\textbf{Performance comparison SFT on AIME 2024 and 2025.} Mean Pass@1 accuracy and mean token length per training epoch.}
\label{fig:aime_sft}
\end{center}
\vskip -0.3in
\end{figure}

\subsection{Results on AIME}

Table~\ref{tab:aime-results} displays the accuracy and the mean number of output tokens on AIME 2024 and 2025. The results show that for both benchmarks, our proposed method of 10-epoch SFT followed by RL improves both accuracy and token efficiency over the original models across nearly all model sizes. Interestingly, the accuracy gains from our method become more pronounced as the model size increases. When RL is applied directly to the original model, the 14B model shows improved token efficiency but a degradation in accuracy. However, by first applying 10-epoch SFT, we significantly boost the model's accuracy, and the subsequent RL phase further enhances it. The 7B model exhibits a similar trend, with accuracy improving incrementally first with extensive SFT and then with RL. In contrast, for the 1.5B model, the 10-epoch SFT did not yield a substantial accuracy increase.

\subsection{Results on MATH-500}

Table~\ref{tab:pass1_tokenlen} presents the results on MATH-500, a benchmark considered less difficult than AIME. The lower number of tokens used for inference suggests that MATH-500 has a different difficulty profile. Despite this difference, we observe a trend similar to the AIME results when applying our training recipe. The 10-epoch SFT phase increases both accuracy and mean token length, while the subsequent RL phase successfully improves accuracy further while reducing the token length. This demonstrates that our method can enhance accuracy while managing token usage, even on benchmarks with different difficulty levels and token requirements for inference.

\subsection{The Impact of Extensive SFT}

As shown in Table~\ref{tab:aime-results}, applying RL within our framework helps reduce the number of tokens required for inference. However, for larger models like the 14B variant, applying RL alone can be unstable, leading to a drop in accuracy. Indeed, prior work has reported the necessity of an initial SFT phase to stabilize the RL process \citep{deepseekai2025deepseekr1incentivizingreasoningcapability}. We investigated how much SFT is necessary before RL when building a specialized mathematical model. As shown in Figure~\ref{fig:aime_sft}, there is a clear trend where accuracy improves as the number of SFT epochs increases. Interestingly, training for only one epoch significantly increases the average token length but leads to a sharp drop in accuracy. These findings suggest that for creating a specialized math model, a prolonged SFT phase is crucial for enabling stable and effective subsequent RL training.

\begin{figure}[t]
\begin{center}
\centerline{\includegraphics[width=\columnwidth]{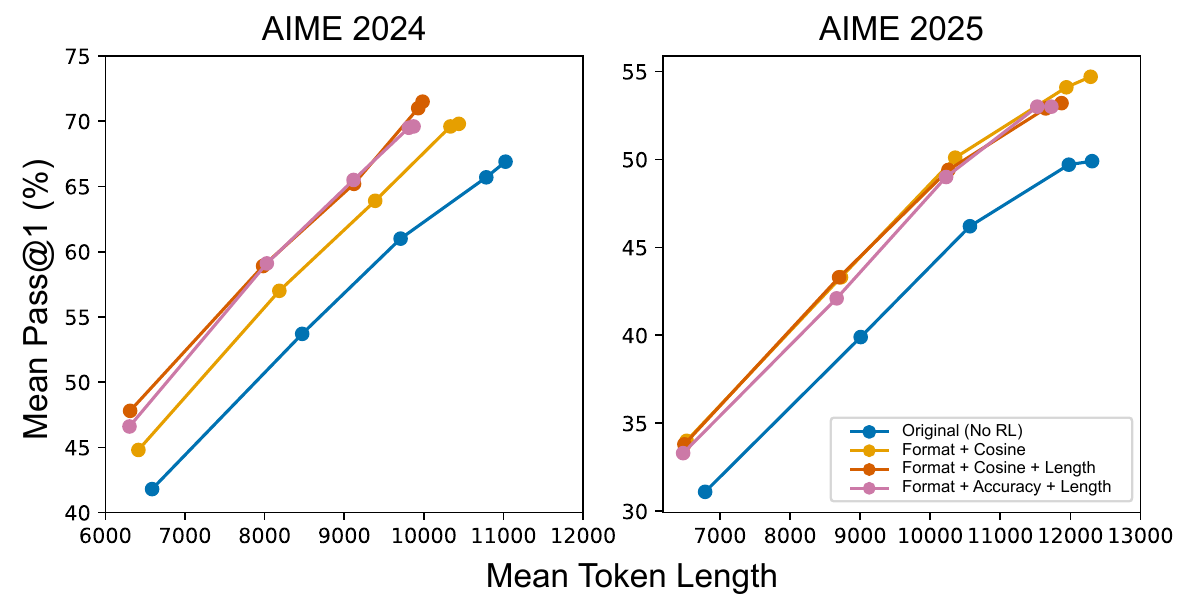}}
\caption{\textbf{Ablation study of Reward functions.} The mean Pass@1 accuracy versus the mean token length for different combinations of reward functions. To clearly illustrate the performance at different token budgets, points are also plotted for outputs truncated at maximum token lengths of 8k, 12k, 16k, 24k, and 32k.}
\label{fig:reward}
\end{center}
\vskip -0.3in
\end{figure}

\subsection{Ablation on Reward Functions}

We investigated how the combination of the reward functions for RL affects our model's performance. To test this, For this analysis, we evaluated three distinct reward configurations. The \textit{accuracy reward} is a binary reward for correct answers~\cite{deepseekai2025deepseekr1incentivizingreasoningcapability}, while the \textit{length penalty} penalizes longer solutions. The \textit{cosine reward}~\citep{yeo2025demystifying} provides a reward based on both accuracy and solution length. A \textit{format reward} was used in all experiments to ensure proper output structure. To clearly illustrate the trade-off between inference tokens and accuracy, we also plot accuracy at different token budget cutoffs. Figure~\ref{fig:reward} shows that incorporating a length penalty effectively reduces the average number of tokens required for inference. When comparing the accuracy-based rewards, the cosine reward yields slightly higher accuracy than the binary accuracy reward alone. In our final training recipe, we adopted a combination of the cosine and length penalty to strike a balance between token efficiency and accuracy.

\begin{figure}[t]
\begin{center}
\centerline{\includegraphics[width=\columnwidth]{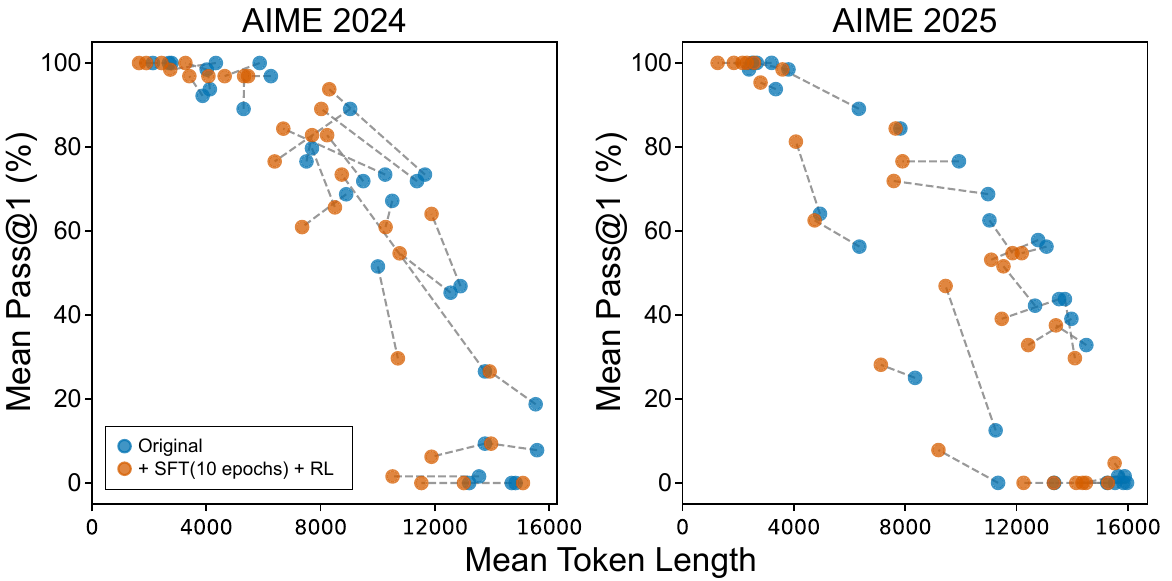}}
\caption{\textbf{Per-problem changes in mean pass@1 and token length from the original model to our proposed recipe.} 
This plot illustrates the shift in performance and efficiency for each problem after applying our training recipe.}
\label{fig:per_problem}
\end{center}
\vskip -0.3in
\end{figure}

\subsection{Analysis of Per-Problem Performance}

To gain a more granular understanding of our recipe's impact, we conducted a per-problem analysis of its effects on both accuracy and token efficiency. For this experiment, we evaluated the performance of our final model against the original DeepSeek-R1-Distill-Qwen-14B baseline on every problem in the AIME 2024 and AIME 2025 benchmarks. As illustrated in Figure~\ref{fig:per_problem}, the results demonstrate a clear and consistent improvement across the majority of problems. For most questions, our recipe not only enhances the mean pass@1 score but also substantially reduces the solution length, indicating a significant gain in overall efficiency. For problems where the baseline model already achieved a high accuracy (i.e., approaching $100\%$), our method successfully maintains or further improves accuracy without any instances of performance degradation. In the mid-range of difficulty, where the accuracy was between $10\%$ and $80\%$, our model demonstrates its most significant impact. While a few problems show minor regressions, the overwhelming trend is a simultaneous improvement in both accuracy and token efficiency. However, for the most challenging problems where the accuracy was initially low, our recipe shows limited gains in improving its performance. This suggests that improving performance on the most difficult problems remains a key challenge for future work.

\subsection{Final Performance on the AIMO Benchmark}

Finally, we evaluated our proposed recipe on the AIMO benchmark, which is both highly challenging and completely leak-free, to assess its real-world performance. Since the number of evaluation in the AIMO competition is strictly limited, we assess the performance of our recipe by its final ranking in the competition. Our model achieved a score of $29/50$ on the public set (equivalent to 4th place) and $28/50$ on the private set (equivalent to 8th place) out of $2,212$ competing teams. Achieving a consistently high score on both the public and private sets, amidst a large number of participants, demonstrates that our method is robust and capable of delivering genuinely high performance.

\section{Conclusion}
\label{conclusion}

In this work, we proposed and validated a practical training recipe that synergistically combines Supervised Fine-Tuning (SFT) and Reinforcement Learning (RL) with GRPO to advance the mathematical reasoning of LLMs. Our core finding is that these two methods play complementary roles: an extended SFT phase is crucial for pushing the model's accuracy to its limits, while a subsequent GRPO phase dramatically improves token efficiency without compromising this peak performance. This sequential strategy moves beyond viewing SFT and RL as competing alternatives, establishing a clear, effective pathway to developing models that are both highly accurate and efficient.

The efficacy of our approach was rigorously demonstrated through top-tier performance on challenging benchmarks, including AIME and MATH. Most notably, our model achieved a high rank in the AI Mathematical Olympiad (AIMO), a highly competitive and strictly leak-free competition, confirming the robustness and real-world effectiveness of our method. These results provide a battle-tested blueprint for the community, showcasing a reliable method to build the next generation of state-of-the-art mathematical reasoners that balance exceptional performance with practical applicability.

\nocite{langley00}

\bibliography{main}
\bibliographystyle{icml2025}


\end{document}